\documentclass[sigconf]{acmart}
\AtBeginDocument{%
  }

\copyrightyear{2026}
\acmYear{2026}
\setcopyright{cc}
\setcctype{by}
\acmConference[MM '26] {Proceedings of the 34th ACM International Conference on Multimedia}{November 10--14, 2026}{Rio de Janeiro, Brazil.}
\acmBooktitle{Proceedings of the 34th ACM International Conference on Multimedia (MM '26), November 10--14, 2026, Rio de Janeiro, Brazil}
\acmISBN{979-8-4007-2213-4/2026/11}
\acmDOI{10.1145/3767308.3835473}
\usepackage{url}

\newcommand{\ourMthd}{DreamStyle3D}

\newcommand{\moduleD}{Decoupled Dual Cross-Attention}

\newcommand{\moduleS}{Style Disentanglement Augmentation}

\newcommand{\tabref}[1]{Tab.~\ref{#1}}
\newcommand{\figref}[1]{Fig.~\ref{#1}}

\newcommand{\secref}[1]{Sec.~\ref{#1}}

\newcommand{\myPara}[1]{\noindent\textbf{#1}}

\begin{document}

\title{\ourMthd{}: Efficient 3D Stylized Asset Generation via Dual-Attention Disentanglement}

\author{Kai Wang}
\affiliation{%
  \institution{VCIP, Nankai University}
  \city{Tianjin}
  \country{China}}
\affiliation{%
  \institution{Zhongguancun Academy}
  \city{Beijing}
  \country{China}}
\email{wangkaink25@mail.nankai.edu.cn}

\author{Ziheng Ouyang}
\affiliation{%
  \institution{VCIP, Nankai University}
  \city{Tianjin}
  \country{China}}
\email{oyzh@mail.nankai.edu.cn}
  
\author{Xuying Zhang}
\authornote{Project Leader. \quad \ding{41} Corresponding author.}

\affiliation{%
  \institution{Joy Future Academy, JD}
  \city{Beijing}
  \country{China}
  }
\affiliation{%
  \institution{VCIP, Nankai University}
  \city{Tianjin}
  \country{China}}
\email{zhangxuying1004@gmail.com}

\author{Ming-Ming Cheng}
\correspondingauthor
\affiliation{%
  \institution{VCIP \& AAIS, Nankai University}
  \city{Tianjin}
  \country{China}}
\email{cmm@nankai.edu.cn}

\author{Qibin Hou}
\affiliation{%
  \institution{VCIP \& AAIS, Nankai University}
  \city{Tianjin}
  \country{China}}
\email{houqb@nankai.edu.cn}

\renewcommand{\shortauthors}{Wang et al.}

\begin{abstract}
With the growth of gaming, animation, and virtual reality industries, the demand for efficient generation of stylized 3D assets is rapidly increasing.
However, existing approaches still struggle to jointly preserve style fidelity, geometric consistency, and generation efficiency, as most of them still rely on indirect 2D-to-3D stylization pipelines.
This motivates a native 3D stylization framework that can explicitly disentangle style from geometry while remaining efficient.
To this end, we propose \ourMthd{}, an efficient framework for stylized 3D asset generation built on a \moduleD{} mechanism.
Our method explicitly separates geometric and stylistic features to enable efficient style injection while preserving structural consistency, and further adopts a lightweight training strategy to enhance style consistency and model generalization.
In addition, we build an automated data pipeline and construct a dataset of about 15K content-style-stylized triplets for training and evaluation.
Extensive experiments demonstrate that our \ourMthd{} can generate high-fidelity, geometrically consistent stylized 3D assets within 10 seconds, substantially improving efficiency while maintaining superior style quality and offering a new solution for 3D content creation.
The project is available at \url{https://github.com/NK-JittorCV/nk-3D/tree/main/models/DreamStyle3D}.
\end{abstract}

\ccsdesc[300]{Computing methodologies~Appearance and texture representations}
\ccsdesc[500]{Computing methodologies~Texturing}
\ccsdesc[100]{Computing methodologies~Learning latent representations}

\keywords{3D Stylization, Decoupled Dual Cross-Attention, Style-Geometry Disentanglement, Lightweight Framework}
\begin{teaserfigure}
  \centering
  \includegraphics[width=0.95\textwidth]{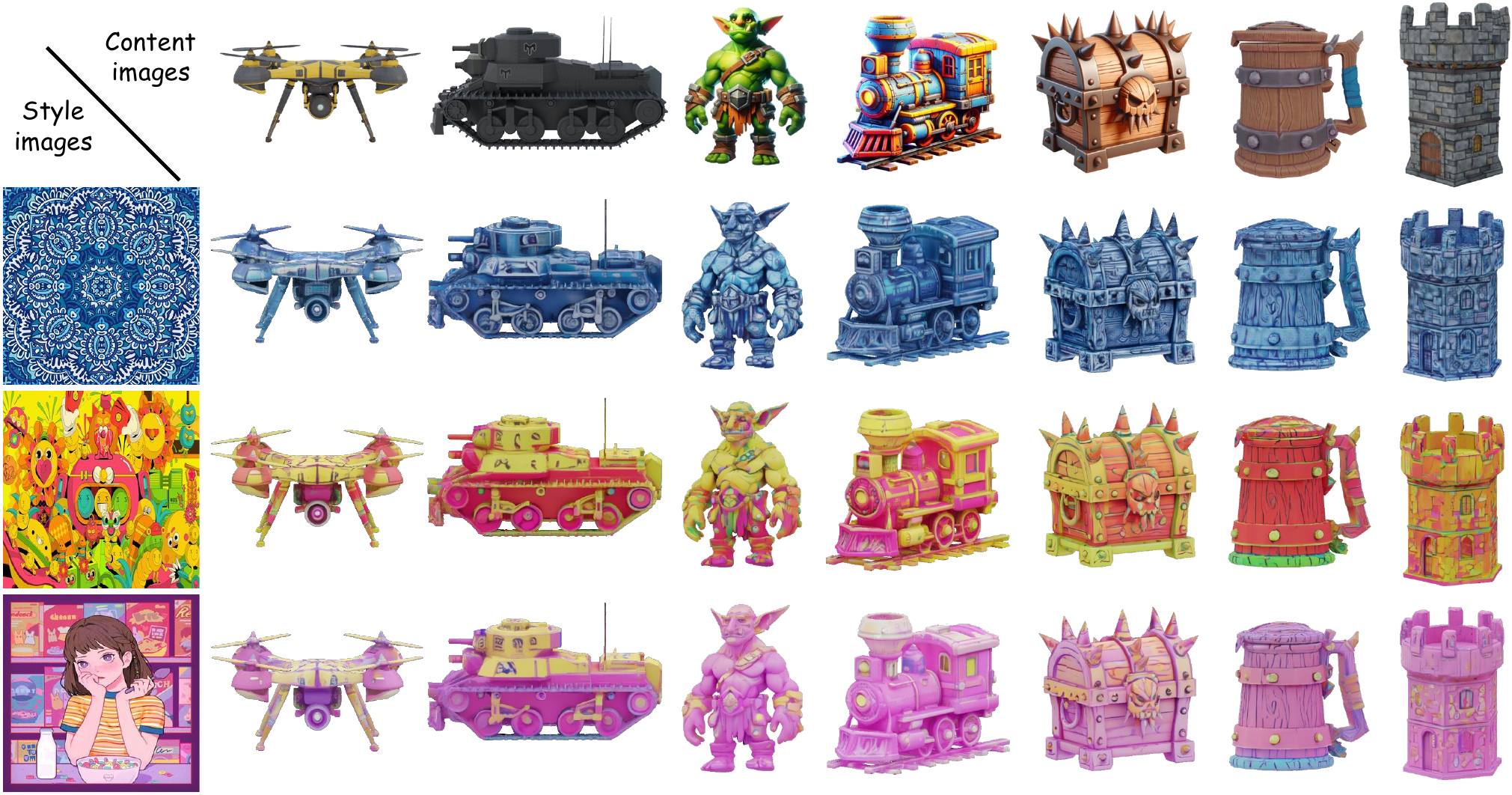}
  \caption{Given a content image and a style image, our \ourMthd{} can generate high-fidelity 3D stylized assets within 10s.}
  \label{fig:teaser}
\end{teaserfigure}


\maketitle

\section{Introduction}

With the rapid development of 3D generative models, automated 3D content generation has opened up unprecedented possibilities for creators~\cite{zhang2024clay,wu2024direct3d,zhang2025tar3d}.
It is noteworthy that in digital industries such as gaming, animation, and virtual reality, the demand for customized 3D asset generation is also steadily increasing. 
Beyond faithful geometry and texture reconstruction, many applications also require controllable stylization, where a 3D object should preserve its structural identity while exhibiting the visual characteristics of a reference style image. 
This makes image-driven 3D stylization an important yet challenging problem, as it requires jointly achieving style fidelity, geometric consistency, and high efficiency.

\begin{figure}[t]
    \centering
    \includegraphics[width=0.90\linewidth]{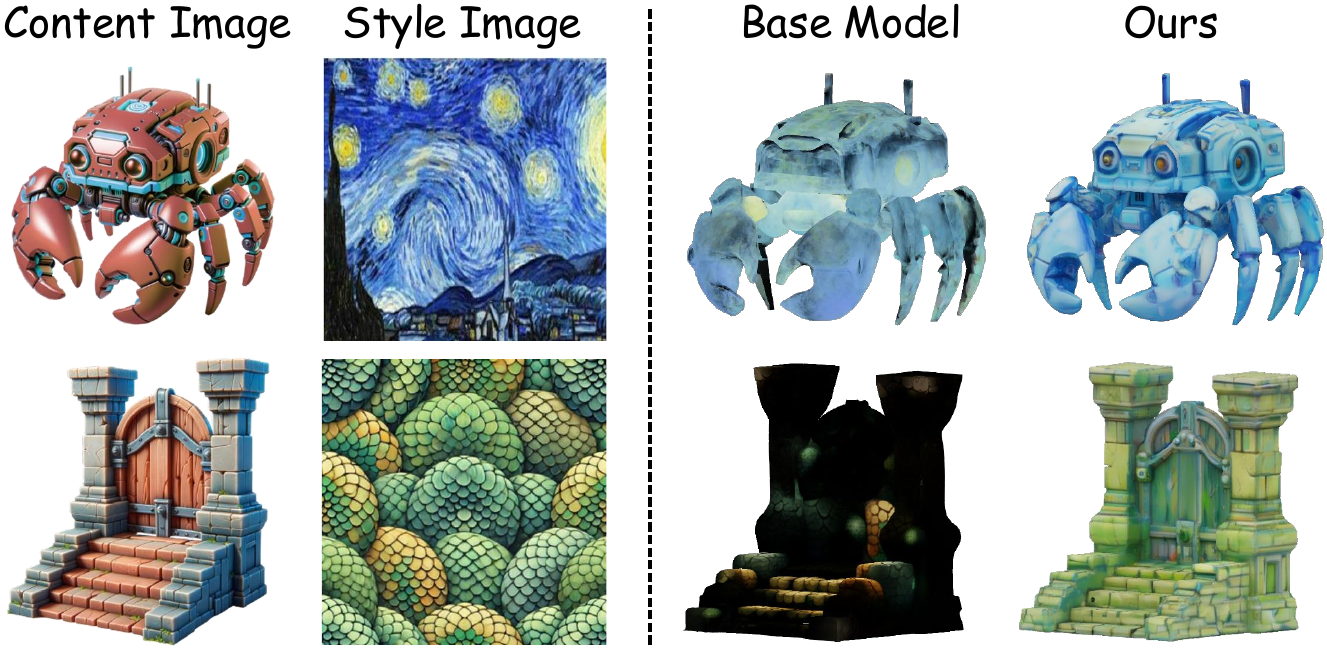}
    \caption{
    Visualization of artifacts caused by the entanglement of style and geometric structure in the base model.
    }
    \label{fig:intro}
\end{figure}

Existing 3D stylization methods mainly fall into two categories: optimization-based methods~\cite{michel2022text2mesh,poole2022dreamfusion,chen2023fantasia3d,zhang2024temo,hunyuan3d2025hunyuan3d} and feature-fusion methods~\cite{oztas20253d}.
Despite their different technical routes, these approaches largely follow an indirect 2D-to-3D stylization paradigm and rely on 2D priors or intermediate representations, making it difficult to jointly preserve style fidelity, geometric consistency, and generation efficiency. 
Particularly, the former methods either perform score distillation sampling (SDS) on 2D pre-trained models for color regression, or employ 2D diffusion models to generate multi-view images mapped onto 3D meshes.
Although visually faithful, these methods require considerable training and inference time.
In contrast, feature-fusion methods inject style information into a large reconstruction model (LRM)~\cite{hong2023lrm} that transfers 2D sparse views into 3D objects via a feed-forward manner.
These methods offer fast generation speed, yet suffer from semantic drift and structural distortion due to the coupled style and geometry.
In this paper, we eschew the previous indirect generation strategy and instead advocate the direct generation of 3D stylized assets.
Inspired by the recent success of 3D structured latents (SLAT)~\cite{xiang2025structured,He_2025_ICCV}, we leverage a native 3D diffusion model to achieve this objective.
However, native 3D stylization is still constrained by two critical bottlenecks: \textbf{1)} Visual artifacts resulting from the entanglement of style and geometry, as shown in Fig.~\ref{fig:intro}; \textbf{2)} Train-test inconsistency caused by the scarcity of 3D stylization datasets.

To address these challenges, we propose \ourMthd{}, a native 3D stylization framework for efficient and coherent generation of stylized 3D assets.
Specifically, we introduce a \moduleD{} into the SLAT-based diffusion model.
In this module, object attention and style attention operate independently. The object attention branch focuses on the geometric features of the content image to preserve geometric details, while the style attention branch is responsible for capturing the style features from the style image. This explicit separation allows for a clear disentanglement of the geometry and style representations.
To improve the robustness and generalization of our method, we design a \moduleS{} strategy.
By applying structure-aware perturbation to the style image, we remove its structural features and shuffle the image blocks, preserving only the style and texture information. For the content image, we perform color remapping to alter its color distribution, reducing reliance on color information. These data-level perturbations ensure that the model learns independent style and geometric features, thereby facilitating better disentanglement of style and geometry.

Regarding the data limitation, it is challenging to directly acquire triplet samples in the form of (content, style, stylized content) from 3D scenes.
In this work, we develop an automated data generation pipeline to tackle this issue.
To be specific, we first feed the content images and the style images into an existing 2D stylization method to generate stylized images, from which we select samples with high visual fidelity.
Based on these samples, we then utilize a pre-trained 3D generation model to obtain stylized 3D objects and subsequently filter out low-quality samples.
Through this data pipeline, we build a stylized 3D asset dataset containing approximately 15K triplets.

We validate the effectiveness of the proposed \ourMthd{} through extensive experiments.
By introducing the native 3D stylization scheme, our \ourMthd{} can generate high-quality stylized objects, as shown in \figref{fig:teaser}, within 10 seconds, achieving a balance between style fidelity and generation efficiency.

Our contributions are summarized as follows:
\begin{itemize}
    \item We propose an efficient stylized 3D asset generation framework, which contains a \moduleD{} to explicitly disentangle and fuse geometric and stylistic representations in native 3D space.
    \item We present a \moduleS{} strategy, combining structure-aware style perturbation and foreground color remapping to improve geometry--style disentanglement and generation stability.
    \item We construct an automated triplet-based data generation pipeline that produces approximately 15K stylized 3D samples, providing standardized paired supervision and benchmarks for model training and evaluation.
\end{itemize}

\begin{figure*}
    \centering
    \includegraphics[width=0.96\linewidth]{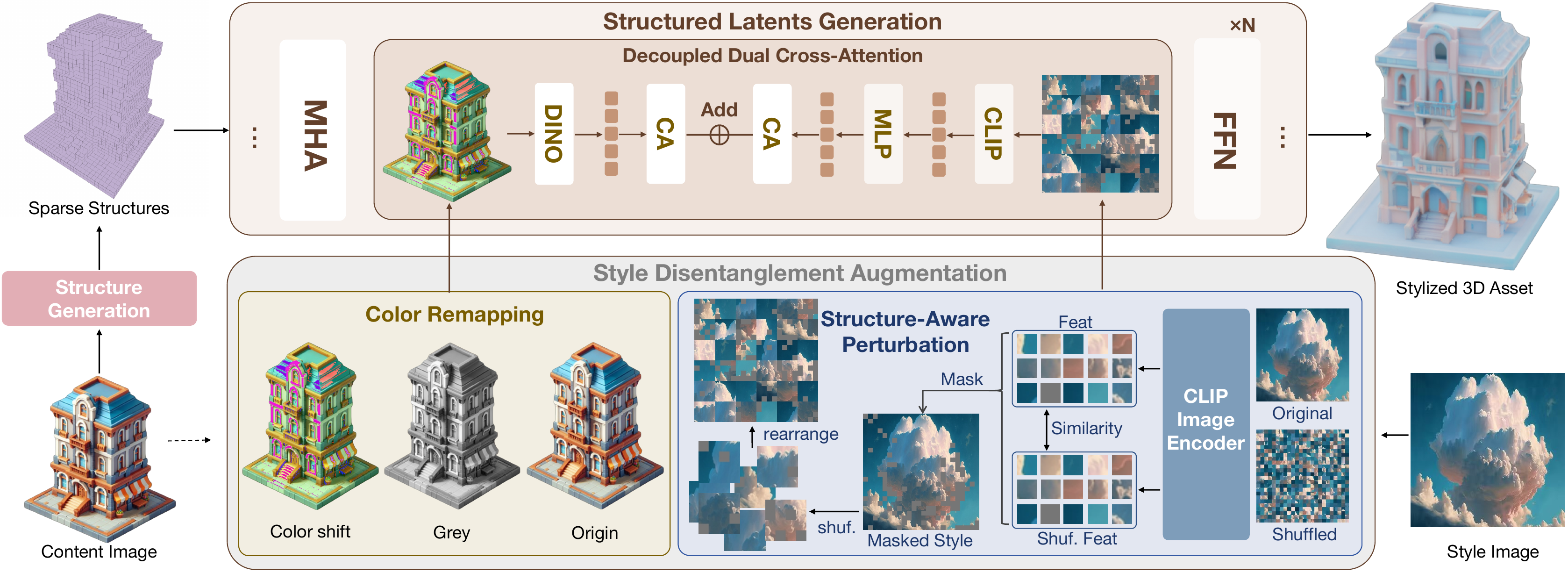}
    \caption{Overall pipeline of the proposed \ourMthd{} approach for stylized 3D asset generation in native 3D space.}
    \label{fig:overall}
\end{figure*}

\section{Related Work}
\myPara{3D Generation.}
Recent years have witnessed significant advances in 3D generation models with the development of diffusion models~\cite{song2020denoising,ho2020denoising,podell2024sdxl,ouyang2026consistency,gong2026direct} and the availability of high-quality 3D datasets~\cite{deitke2023objaverse,stojanov2021using}.
The Score Distillation Sampling method~\cite{poole2022dreamfusion}, which distills 3D information from 2D diffusion models, pioneered the development of diffusion-based 3D generation techniques. 
Then, Zero123~\cite{liu2023zero} addressed the issue of limited views in single-view reconstruction by synthesizing multiple new view images from a single input image through view-conditioned diffusion. 
MV-Dream~\cite{shi2023mvdream} further introduced a multi-view mechanism into diffusion models, enabling the generation of consistent and reliable multi-view images, which provide stable input for downstream 3D asset construction. 
Building upon this, Zero123++~\cite{shi2023zero123++} enhanced the conditional mechanism and training processes, significantly improving the quality and stability of multi-view generation. 
AR123~\cite{zhang2025ar} adopted autoregressive strategies to generate multi-view images from near to far, further enhancing the accuracy and consistency of geometric structures in complex scenes. 
Hunyuan3D~\cite{zhao2025hunyuan3d,hunyuan3d2025hunyuan3d,lai2025hunyuan3d}, driven by large-scale 3D datasets and combining diffusion models with a physically-based rendering framework, significantly improved the overall quality of 3D generation, particularly excelling in texture realism and geometric detail. 
Meanwhile, TRELLIS~\cite{xiang2025structured} introduced native 3D generation based on structured 3D latents, making it possible to perform generation directly in 3D latent space rather than relying solely on intermediate multi-view representations.

\myPara{Style Transfer.}
In the field of stylization, the main research focus lies in the disentanglement of style and content.
Classical 2D neural style transfer methods also study the separation of style and content~\cite{gatys2016image,johnson2016perceptual}, but they typically rely on optimization-based pipelines or models trained from scratch; here we focus on recent lightweight-training methods built on large pretrained image generation models.
Among these recent methods, adapter-based tuning~\cite{hertz2024style,ye2025stylemaster,liu2024ada,wang2024instantstyle,xu2025clgc} and LoRA-based fine-tuning~\cite{frenkel2024implicit,ouyang2025k,chen2025consislora,xu2026smrabooth,xu2026disco} have become two representative paradigms. 

Adapter-based approaches typically rely on attention mechanisms to achieve style consistency. Representative works include StyleAligned~\cite{hertz2024style}, which shares self-attention and uses AdaIN~\cite{huang2017arbitrary} to align the query and key features of the target and reference images; StyleAdapter~\cite{wang2023styleadapter}, which reduces semantic interference by removing class tokens and shuffling positional embeddings; and CSGO~\cite{xing2024csgo}, which achieves disentanglement of object and style through multi-attention injection. 
LoRA-based style transfer methods can be roughly divided into two branches. One branch is related to concept-level personalization, a setting popularized by DreamBooth~\cite{ruiz2023dreambooth}, with later works such as ZipLoRA~\cite{shah2024ziplora}, B-LoRA~\cite{frenkel2024implicit}, AgeBooth~\cite{zhu2025agebooth}, and K-LoRA~\cite{ouyang2025k} further improving disentanglement or fusion through LoRA weight analysis and composition.
The second branch focuses on data-driven LoRA fine-tuning, where the base model learns style transfer paradigms from constructed datasets, such as OmniStyle~\cite{wang2025omnistyle}, USO~\cite{wu2025uso} and StyleExpert~\cite{zhu2026mixture}. 

Inspired by advances in 2D style transfer, recent work has begun to explore generative 3D stylization through image-conditioned pipelines, where stylization must preserve geometric and multi-view consistency in addition to visual style~\cite{ye2024jigsaw3d}.
Style3D~\cite{song2024style3d} directly transfers self-attention features from 2D reference images into multi-view generation;
3DStyleLRM~\cite{oztas20253d} fuses style information via a linear combination of CLIP-based~\cite{radford2021learning} reference features; 
and StyleTex~\cite{xie2024styletex} adopts orthogonal projection in a semantically-aware feature space to decompose style diffusion loss, though its rendering optimization at test time is computationally expensive. 

However, these methods still rely on indirect 2D-to-3D generation pipelines.
We instead propose a native 3D stylization strategy that performs style transfer directly in native 3D latent space.

\section{Methodology}

\subsection{Preliminaries}\label{subsec:trellis}
We introduce the preliminaries of TRELLIS~\cite{xiang2025structured}, the base native 3D model adopted in this work, which is beneficial for understanding the designs of our \ourMthd{}.

\myPara{Structured Latent Representation.} For a 3D mesh, the geometry and appearance information are encoded as a structured latent representation (SLAT), denoted as \(z\), integrating a sparsely-populated 3D grid with local latents:
\begin{equation}
\mathbf{z} = \{(z_i, p_i)\}_{i=1}^{L}, z_i \in \mathbb{R}^C, p_i \in \{0,1,\ldots, N_{r}-1\}^3,
\end{equation}
where \(N_{r}\) is the resolution of the 3D grid, \(L\) is the number of active voxels occupied by the mesh, \(p_i\) is the coordinate of an active voxel, and \(z_i\) is the corresponding local feature.

\myPara{Sparse Structure Generation.}
The model first performs rectified flow-based diffusion sampling over a \(N_{r}^3\) dense grid to predict occupancy probabilities.
They are further converted to active voxels, which form a sparse structure to capture the global topology and spatial layout of the object.

\myPara{Structured Latent Generation \& Versatile Outputs.}
Given the sparse scaffold and conditional image, a diffusion model based on a sparse flow transformer samples the local latents associated with the active voxels.
Then, these structured latents can be decoded into diverse 3D representations, such as 3D Gaussians and meshes.

\subsection{\ourMthd{}}\label{subsec:ourmthd}

Existing 3D stylization methods, which follow the indirect 2D-to-3D stylization scheme for texturing 3D assets, struggle to balance stylistic fidelity and generation efficiency.
In this work, we shift to a direct stylized asset generation scheme in native 3D space.
\figref{fig:overall} illustrates the \ourMthd{} pipeline for generating stylized 3D assets.
Specifically, we first feed the content image to the sparse structure generator of the base model to generate a coarse geometry consisting of active voxels.
Then, we integrate the style image into the structured latent generator to generate SLAT representations, which are decoded as a mesh that reflects the topological structure and stylized Gaussian points used for texture baking.
However, the incorporation of style images can lead to the entanglement of stylistic and geometric information.
To solve this problem, we develop a \moduleD{} to explicitly separate geometric and stylistic representations, as shown in \secref{subsec:attn}.
In addition, we design a \moduleS{} strategy to reduce style interference in the content image and structural feature interference in the style image, 
as detailed in \secref{subsec:disentangling}.

\subsection{\moduleD{}}\label{subsec:attn}
The base 3D model leverages a pre-trained DINO~\cite{oquab2023dinov2} to extract features from the content image, which are fused with the structured latent representation via cross-attention to generate detailed appearance.  
However, this attention mechanism primarily reconstructs geometric details (e.g., edges and corners) and lacks the capacity to capture global semantic or stylistic patterns.  

Empirical studies indicate that DINO is trained under a self-distillation objective, leading to an inherent bias towards local spatial structures rather than visual style.  
As a result, the original geometry-guided attention is insufficient for controllable image style transfer.
In contrast, CLIP~\cite{radford2021learning} is pretrained on large-scale image–text pairs,  providing strong semantic abstraction and stylistic separability in its feature space.  
This allows CLIP to encode both semantic context and global artistic traits,  making it suitable as a complementary style encoder.

Our design is motivated by the observation that geometry and stylistic appearance are distinct modalities, and directly merging their features in a shared attention pathway can easily entangle structural and stylistic cues.
A straightforward idea to incorporate style features is to concatenate style features with geometric features and feed them into a shared cross-attention layer.
However, it is prone to resulting in visual artifacts, as shown in \figref{fig:intro}.

Instead, we develop a \moduleD{} module, in which the cross-attention layers for geometric features and style features are separate.
Specifically, this module is inserted into the cross-attention layers of the structured latent generator, where the latent tokens interact with geometry and style features through two decoupled branches.
In this module, the content image is encoded by DINO to obtain a structural feature map \(F_c\). 
Meanwhile, CLIP is employed to extract high-level style features \(F_s\), which are projected through a trainable two-layer MLP into the same embedding space as \(F_c\), yielding aligned style representations \(F_s'\).

Subsequently, this module transforms the structured latent tokens \(Z\) into query vectors \(Q = ZW_q\), and performs two parallel attention computations using the DINO-derived geometric features and the CLIP-derived style features as separate key–value sets:
\begin{gather}
A_c = \mathrm{Softmax}\!\left(\frac{QK_c^\top}{\sqrt{d}}\right)V_c, 
A_s = \mathrm{Softmax}\!\left(\frac{QK_s^{\top}}{\sqrt{d}}\right)V_s,
\end{gather}
where \(K_c = F_cW_k, V_c = F_cW_v\),  
and \(K_s = F_s'W_k', V_s = F_s'W_v'\) are the key and value matrices from geometric features and style features, respectively.
We perform element-wise addition on the outputs of the separated attention to obtain the joint representation $F_{\text{out}}$.

Finally, the decoder reconstructs 3D assets that maintain geometric fidelity while exhibiting coherent visual style.

Note that in this module, the two-layer style projection MLP together with \(W_k'\) and \(W_v'\) are trainable, while the remaining weights are frozen.
This lightweight design enables \moduleD{} to improve style expressiveness and visual coherence without modifying the backbone network.

\subsection{Style Disentanglement Augmentation}\label{subsec:disentangling}
Although the proposed \moduleD{} enables the disentangled fusion of geometry and style within the structured latent space, the model may still capture correlated color or local structural cues from both the content and style images during training,  
leading to residual semantic coupling.  
To further enhance the model’s ability to distinguish between geometry and style, we propose a \moduleS{} strategy that explicitly perturbs both content and style inputs at the data preprocessing stage. This augmentation introduces distributional randomness that prevents color–structure co-adaptation  and improves the purity of learned style representations.

\myPara{Structure-Aware Perturbation for Style Images.}
Style reference images often contain salient object boundaries or spatial layouts,  
which may bias the model toward reconstructing geometric structures rather than learning pure stylistic representations.  
To alleviate this issue, we propose a structure-aware perturbation strategy that suppresses structure-sensitive cues while preserving style-dominant local patterns, encouraging the network to focus on texture and brushstroke statistics rather than spatial layout.

Our key observation is that spatial shuffling disrupts the global arrangement of the image while largely preserving local color and texture statistics within each patch. 
Consequently, patches that remain similar before and after shuffling are more likely to encode texture-dominant cues, whereas structure-dependent regions tend to become unstable under spatial disruption. Based on this observation, we identify texture-stable regions and suppress structure-sensitive ones for style learning.

Specifically, we divide the style image $x$ into non-overlapping $14\times14$ patches, and generate a randomly shuffled version of it, denoted as $x_{shuf}$. 
We then extract patch features using the CLIP vision encoder and compute a patch-wise similarity matrix:
\begin{equation}
S_{ij} = \frac{f_i(x) \cdot f_j(x_{shuf})}{\|f_i(x)\|\|f_j(x_{shuf})\|},
\end{equation}
where $f_i(\cdot)$ denotes the feature of the $i$-th patch.
Based on the similarity scores, we preserve the top-$r$ proportion of high-similarity regions to construct a binary mask $m$, which retains texture-stable regions while suppressing structure-sensitive ones.
The mask is applied back to the style image, preserving texture-stable patches while filling the remaining regions with their local mean:
\begin{equation}
x_{mask} = m \odot x + (1 - m) \odot \bar{x}.
\end{equation}
Finally, a second shuffling with a larger patch size is applied to further remove any residual structural alignment.  

During training, we employ the full mask+shuffle perturbation,  
while during inference, only the large-scale shuffling is performed to maintain global style consistency.  

\begin{figure}[t]
    \centering
    \includegraphics[width=0.95\linewidth]{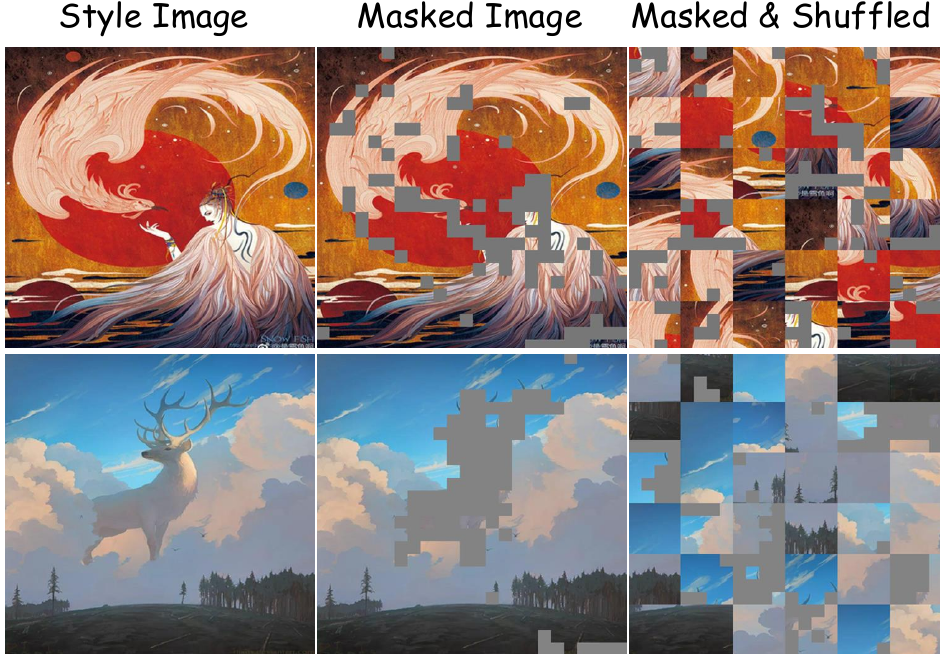}
    \caption{Visualizing Style Disentanglement Augmentation.}
    \label{fig:disentangling}
\end{figure}

As shown in Fig.~\ref{fig:disentangling}, after the masking and shuffling operations, the structural information in the image is significantly reduced, while the texture and style components are preserved. 
This strategy exploits the fact that structure-dependent regions exhibit significant similarity decay after shuffling, whereas texture-dominant regions remain relatively stable, thereby achieving adaptive disentanglement between structure and style. 
Consequently, the model learns purer and spatially invariant style representations.

\myPara{Color Remapping for Content Images.}
While content images primarily convey geometry and structure, their color distributions may also provide spurious cues that interfere with style disentanglement.  
To address this issue, we introduce a foreground color remapping strategy that stochastically weakens foreground chromatic cues while preserving geometric structure, thereby encouraging the model to rely more on contours and shape information.

To reduce color reliance without altering the overall geometric layout, we apply the remapping to the foreground region while keeping the background unchanged.
Given an image $x \in [0,1]^{3\times H\times W}$ and its foreground mask $m$, the unified operator is defined as:
\begin{equation}
A(x,m) = \Phi(x)\odot m + x\odot(1-m),
\end{equation}
where $\odot$ denotes element-wise multiplication,  
and $\Phi$ is a stochastic color transform applied only to the foreground.
With probabilities $(p_{\text{orig}}, p_{\text{gray}}, 1-p_{\text{orig}}-p_{\text{gray}})$,  
$\Phi$ either keeps the foreground unchanged as an identity branch, converts it to grayscale,  
or perturbs its color distribution by quantizing RGB values into $L$ bins  
and cyclically shifting them by an integer offset $\Delta \!\sim\! \mathcal{U}([-S,S]^3)$:
\begin{equation}
\Phi(x) = \frac{((\mathrm{Round}(x\times 255 / L) + \Delta) \bmod 256)}{255}.
\end{equation}

During training, this augmentation produces color-degraded but geometrically consistent samples, forcing the model to rely more on contours and structural cues rather than chromatic shortcuts. 
During inference, it is disabled to preserve the natural appearance of the generated results.

The two augmentations are jointly applied during training in a complementary manner: the content image is regularized to suppress chromatic shortcuts and retain geometry-related cues, while the style image is perturbed to suppress structure-sensitive information and preserve texture-dominant style statistics. 
Together, they explicitly disentangle geometry and style at the data level without introducing additional parameters or loss terms, and they integrate seamlessly with the \moduleD{} module.

\begin{figure*}
    \centering
    \includegraphics[width=1\linewidth]{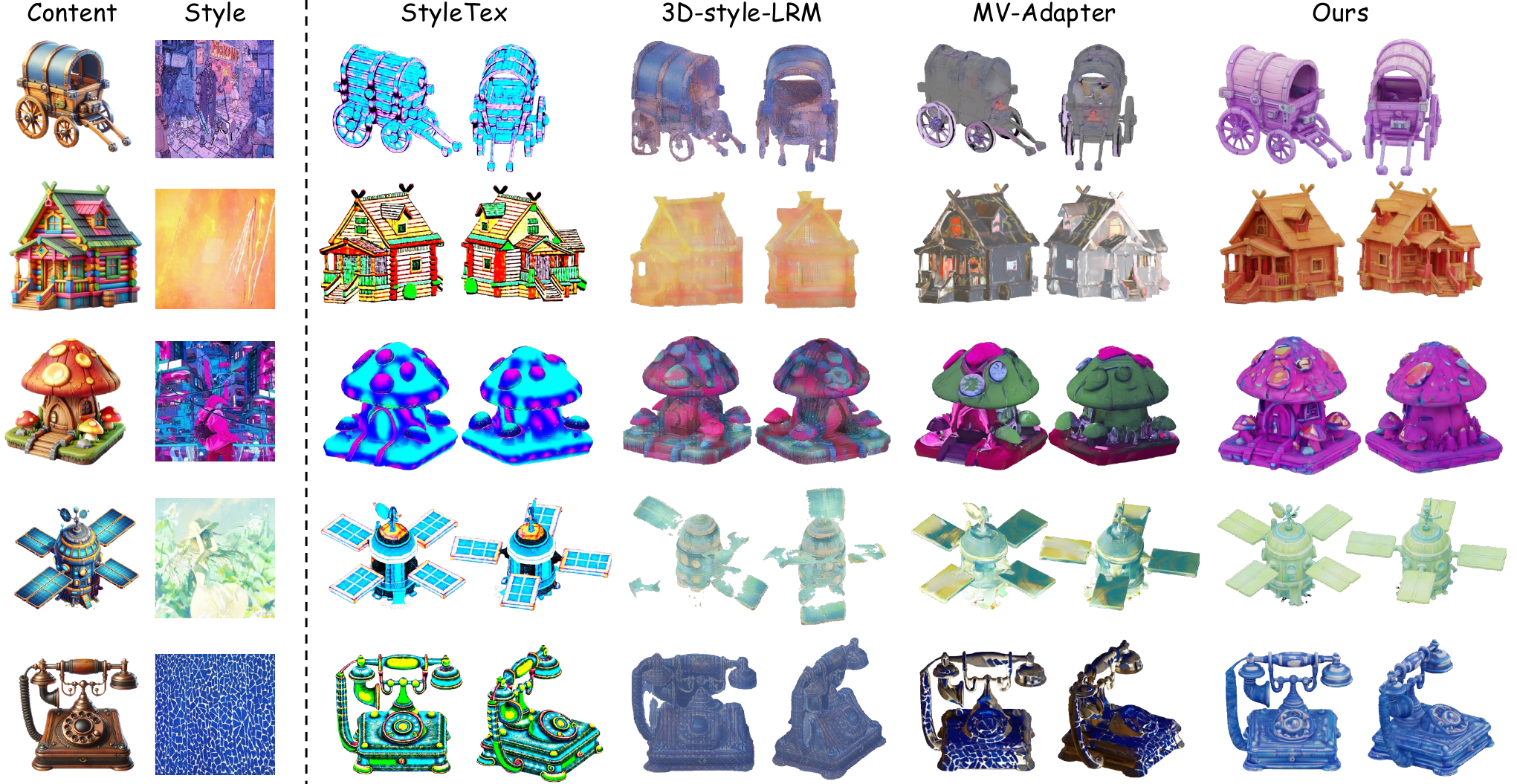}
    \caption{Qualitative comparisons between our \ourMthd{} and recent cutting-edge 3D style transfer methods.}
    \label{fig:Qualitative1}
\end{figure*}

\section{3D Stylization Dataset Curation}\label{sec:dataset}

There is currently a lack of large-scale, paired 3D stylization datasets, which constrains both the end-to-end training and quantitative evaluation of 3D style transfer models.  
To overcome this limitation, we construct an automated 3D stylization dataset curation pipeline, following the triplet form of “content–style–stylized”, integrating both 2D and 3D resources.

Specifically, we select single-object models from Objaverse~\cite{deitke2023objaverse} and Toys4K~\cite{stojanov2021using} and render orthographic-view images as content inputs. 
Each content image is randomly paired with 100 style exemplars sampled from the OmniStyle-150K dataset~\cite{wang2025omnistyle}, forming an initial pool of candidate pairs.
Each pair is processed by the OmniStyle~\cite{wang2025omnistyle} model to generate stylized 2D images, which are further lifted into 3D using the Hunyuan3D-2.1~\cite{hunyuan3d2025hunyuan3d} model, combining geometry reconstruction with style-conditioned texture synthesis.
The resulting dataset contains approximately 15K content-style-stylized asset triplets,   
featuring automated construction, diverse style coverage, and semantic alignment across 2D and 3D domains.

To facilitate the development of 3D stylization, we provide a predefined train/test split for this dataset, where 11K triplet samples are utilized for training while the remaining 4K samples are used for testing.
In practice, we randomly selected 100 samples from the test split and 10 samples with rich details from the Toys4K dataset~\cite{stojanov2021using} to evaluate the performance of the proposed method.
All selected content images in the evaluation are distinct from those used during training to ensure the fairness of the evaluation.

\section{Experiments}
\subsection{Experimental Setup}\label{sec:exp_setting}

\myPara{Training.}
Our training procedure follows the standard latent generation pipeline and fine-tunes only the latent refinement network responsible for local feature denoising.  
The DINO~\cite{oquab2023dinov2} encoder, CLIP~\cite{radford2021learning} encoder, and main Transformer layers are kept frozen, while the parameters of the style attention branch and its MLP projection layers are optimized.
We employ the same Conditional Flow Matching (CFM) objective~\cite{lipman2022flow} as TRELLIS, performing denoising-based learning on the set of local latent variables \(\{z_i\}_{i=1}^{L}\).  
Given a perturbed latent \(z_i(t)\), the model predicts a time-dependent velocity field \(v_\theta(z_i, t)\)  
by minimizing:
\begin{equation}
\mathcal{L}_{\text{CFM}} = \mathbb{E}_{t,z_0,\epsilon}\!\left[\|v_\theta(z,t) - (\epsilon - z_0)\|_2^2\right],
\end{equation}
where \(z_0\) denotes the ground-truth latent and \(\epsilon\) represents Gaussian noise.
During training, the \moduleD{} module injects style-aware features into the vector field prediction,  
enabling joint modeling of geometric fidelity and stylistic control.

\myPara{Implementation Details.}
Our method is fine-tuned based on the TRELLIS framework. For each stylized 3D asset, we rendered 150 images and extracted features to generate structural latent variables as supervisory signals.
We adopted Classifier-Free Guidance(CFG)~\cite{ho2022classifier} with a 0.1 probability of dropping both content and style images simultaneously.
The optimizer used is AdamW~\cite{loshchilov2017decoupled} with a learning rate of $1 \times 10^{-4}$, and the training is performed on a single A800 GPU (80GB) for 130,000 steps with a batch size of 16. During inference, the CFG strength is set to 5, and the sampling steps are set to 50.

\myPara{Baseline Methods.}
We compare \ourMthd{} with existing 3D style generation methods.
The baselines include: 
3D-style-LRM~\cite{oztas20253d}, MV-Adapter~\cite{huang2025mv}, StyleTex~\cite{xie2024styletex}, as well as Hunyuan3D-2.1~\cite{hunyuan3d2025hunyuan3d} and TRELLIS~\cite{xiang2025structured} for evaluating the effectiveness of our framework.
For StyleTex and MV-Adapter, we input the content image into TRELLIS and remove the generated textures from the results, serving as the required object meshes. Simultaneously, we use Qwen3-VL-4B-Instruct~\cite{qwen3technicalreport} to generate the text prompts required for StyleTex.
For Hunyuan3D, the object generation stage uses the content image as input, and the texture generation stage uses the style image as input.
For TRELLIS, during the Sparse Structure Generation stage, we input the content image, and during the Local Latent Generation stage, we tested two different inputs: one with the style image and the other with the concatenation of the content and style images.
All baseline methods are configured with their default settings.

\myPara{Evaluation.}
We render the generated 3D assets, 
as well as the corresponding stylized 3D assets in the test set, from the same six viewpoints for evaluation. 
The rendered images of the stylized 3D assets from the test set are used as reference renders for supplementary reference-based evaluation.
We first evaluate geometry preservation and style consistency using source-image-based metrics together with a VLM-based automatic assessment. 

For geometry preservation, we compute CLIP cosine similarity and DINO cosine similarity between each rendered view and the original content image; higher scores indicate better preservation of object semantics and structure. 
For style consistency, we compute Gram Matrix Similarity and AdaIN Distance between each rendered view and the style reference image; these metrics characterize the agreement with the reference style from the perspectives of style statistics and feature-distribution similarity, respectively.

To obtain a more holistic automatic evaluation from semantic and perceptual perspectives, we further employ Qwen3.5~\cite{qwen3.5} as the vision-language evaluator. 
Specifically, we provide the model with the content reference image, the style reference image, and the six rendered views in a fixed order, and ask it to score each rendered view on three criteria: content preservation, style alignment, and visual quality. Each score is given on a 0--100 scale. We then aggregate these three dimensions and average the resulting scores over the six viewpoints to obtain the final VLM Score.

We further report reference-based metrics as supplementary indicators by comparing the rendered images of generated results with the rendered ground-truth stylized 3D assets from the test set. 
Specifically, we compute CLIP similarity, PSNR, SSIM~\cite{wang2004image}, and LPIPS~\cite{zhang2018unreasonable}, using the ground-truth rendered images as references. 
These metrics provide additional evidence for result quality from the perspectives of high-level semantic consistency, signal fidelity, structural similarity, and perceptual difference. 
All results are averaged over the six viewpoints to provide a comprehensive evaluation of geometric fidelity, style consistency, and visual quality.

\begin{figure*}
    \centering
    \setlength{\abovecaptionskip}{2pt}
    \includegraphics[width=1\linewidth]{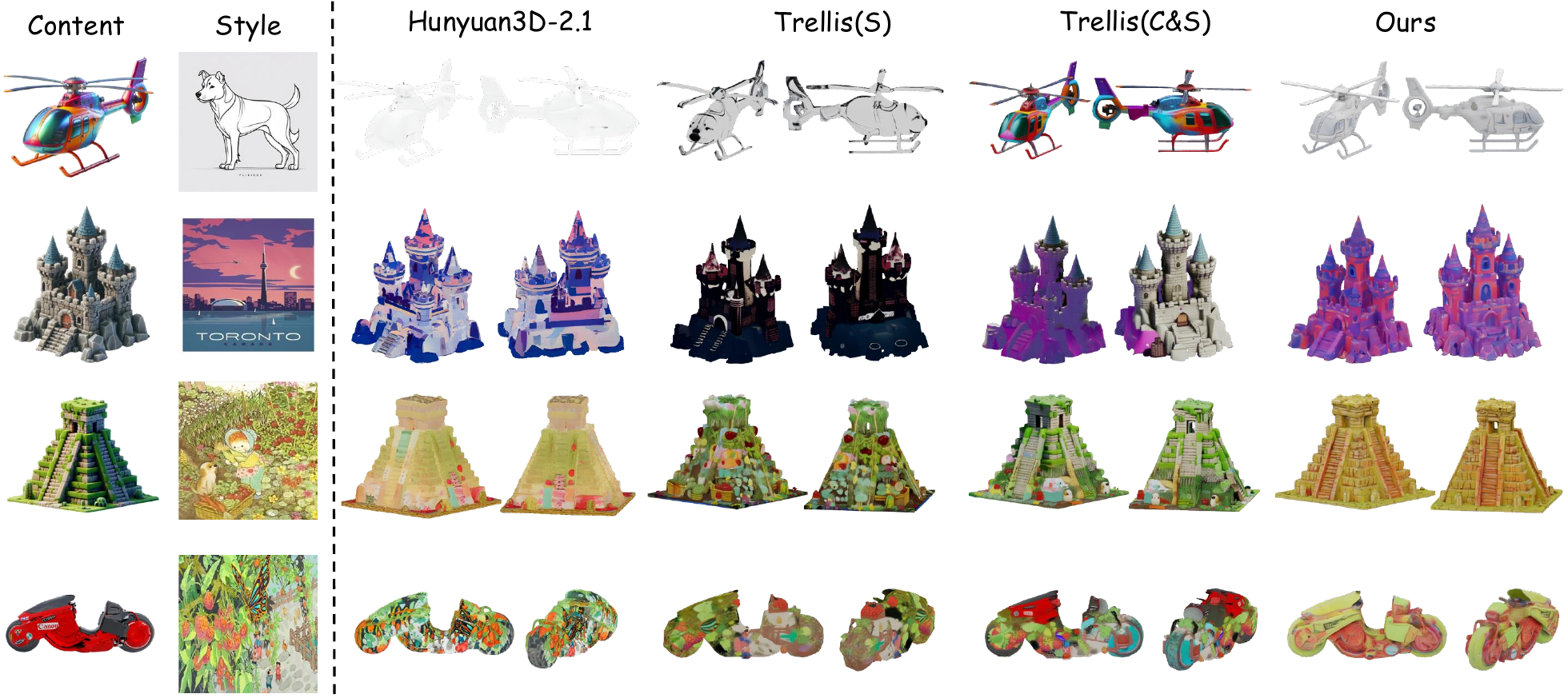}
    \caption{Qualitative comparison between our \ourMthd{} and existing large 3D generative models.}
    \label{fig:Qualitative2}
\end{figure*}

\begin{table}[tp]
    \setlength\tabcolsep{2.5pt}
    \centering
    \small
    \caption{Comparison of 3D style transfer methods using source-image-based metrics, VLM Score, and inference time.}
      \begin{tabular}{lcccccc} 
          \toprule
          Model & CLIP~$\uparrow$ & DINO~$\uparrow$ & Gram~$\downarrow$ & AdaIN~$\downarrow$ & VLM Score~$\uparrow$ & Time   \\ \midrule
          StyleTex & 78.726 & 50.773 & 0.560 & 91.805 & 60.532 & 16min   \\
          3D-style-LRM & 76.559 & 37.364 & 0.529 & 96.254 & 46.829 & $\sim$ 30s  \\
          MV-Adapter & 76.782 & 49.535 & 0.531 & 91.886 & 57.012& $\sim$ 30s  \\
          Ours & \textbf{80.096} & \textbf{63.319} & \textbf{0.516} & \textbf{91.682} & \textbf{73.096} & $\sim$ 10s   \\
          \bottomrule
      \end{tabular}
    \label{tab:quan}
\end{table}
\subsection{Qualitative Results}

\myPara{Comparison with Existing 3D Style Transfer Methods.}
We conducted a qualitative comparison of \ourMthd{} with several existing 3D style transfer methods, and the experimental results are shown in Fig.~\ref{fig:Qualitative1}.
StyleTex~\cite{xie2024styletex} tends to inject style in the form of strong high-frequency surface patterns, which often appear visually detached from the underlying object geometry. 
This suggests that its stylization is biased toward superficial texture transfer, making it difficult to achieve structurally coherent style expression. 
In contrast, 3D-style-LRM~\cite{oztas20253d} captures global style tendency more smoothly, but often weakens local structural expression at the same time. 
Its results generally exhibit over-smoothed colors and missing local components.
MV-Adapter produces more visually plausible stylization in some cases, but its style representation is less stable across different object regions. 
The rendered results frequently show inconsistent color distribution and incomplete texture expression, leading to limited coherence across different views.

Compared with these methods, \ourMthd{} achieves a better balance between style transfer and structural preservation.
The experimental results in Fig.~\ref{fig:Qualitative1} demonstrate that our method  generates high-quality stylized 3D assets across multiple samples while maintaining high style fidelity and geometric consistency.
The results in Fig.~\ref{fig:Qualitative1} show that our method is better able to transfer the target style without introducing severe artifacts, oversmoothing local structures, or causing unstable texture realization. Overall, our method produces stylized 3D assets with stronger structural fidelity and more reliable style consistency than existing approaches.

\myPara{Comparison with 3D Large Models.}
We conducted a qualitative comparison of DreamStyle3D with several existing 3D large model methods, and the experimental results are shown in Fig.~\ref{fig:Qualitative2}.
Current 3D large models do not have the capability to directly generate stylized 3D assets.
The 3D assets generated by Hunyuan3D-2.1 often map texture maps directly onto the objects rather than learning the target style from the style image.
Furthermore, Hunyuan3D-2.1 is not sensitive to certain styles, failing to generate textures correctly.
When TRELLIS uses only the style image as input, it encounters the issue of directly mapping the texture map onto the 3D assets, resulting in a lack of detail in the generated objects.
When both the content and style images are concatenated as input, the style injection is compromised by the content image, leading to suboptimal results.
In contrast, \ourMthd{} accurately extracts the target style from the style image while preserving the geometric details of the object, generating high-quality stylized 3D assets. 

\begin{table}[tp]
    \setlength\tabcolsep{8pt}
    \centering
    \small
    \caption{
    Reference-based quantitative comparison with existing 3D style transfer methods.
    } 
      \begin{tabular}{lcccc} 
          \toprule
          Model & CLIP~$\uparrow$ & PSNR~$\uparrow$  & SSIM~$\uparrow$  & LPIPS~$\downarrow$   \\ \midrule
          StyleTex & 0.819 & 34.698 & 0.761 & 0.314   \\
          3D-style-LRM &  0.790 & 33.902 & 0.759 &  0.323  \\
          MV-Adapter &  0.821 & 34.744 & 0.776  &  0.271  \\
          Ours & \textbf{0.842} & \textbf{34.838} & \textbf{0.802} & \textbf{0.241}  \\
          \bottomrule
      \end{tabular}
    \label{tab:quan1}
\end{table}

\subsection{Quantitative Results}

\tabref{tab:quan} quantitatively compares \ourMthd{} with existing 3D style transfer methods in terms of source-image-based metrics, VLM Score, and inference time.
For geometry preservation, \ourMthd{} achieves the best CLIP and DINO scores, outperforming the strongest baselines by 1.37 and 12.55, respectively.
\ourMthd{} also performs best on Gram and AdaIN and improves the VLM Score over the strongest baseline by 12.56, demonstrating stronger style consistency and overall perceptual quality.
Beyond generation quality, \ourMthd{} also shows a clear advantage in computational efficiency. 
Compared with StyleTex, which requires about 16 minutes of optimization, our method only takes about 10 seconds, making it approximately 96 times faster. 
It is also about three times faster than MV-Adapter and 3D-style-LRM, highlighting its practical advantage for efficient large-scale stylized 3D generation.

\tabref{tab:quan1} further reports the reference-based quantitative comparison, where the reference images are rendered from the stylized 3D assets in the test split. 
Unlike source-image-based comparisons, this group of metrics directly compares the rendered outputs with corresponding 3D reference renders, and therefore provides a more direct measure of consistency in terms of 3D structure and appearance realization. 
We thus use these metrics as supplementary reference-based evaluation to validate the performance of our method from a more direct 3D reference perspective. 
As shown in the table, \ourMthd{} achieves the best performance on CLIP similarity, PSNR, and SSIM, while obtaining the lowest LPIPS value, and the overall trend remains consistent with the source-image-based metrics and the VLM-based evaluation discussed above.

These results confirm improved 3D consistency and rendering quality, while an additional user study reported in the appendix further shows that \ourMthd{} achieves the highest preference rate.

\subsection{Ablation Analysis}
To validate the effectiveness of \moduleS{}, we conduct an ablation study on its two components: Structure-Aware Perturbation (SAP) for style images and Color Remapping (CR) for content images.
Fig.~\ref{fig:abl} and Tab.~\ref{tab:abl} evaluate SAP and CR, while additional ablations on DDCA are provided in the appendix.
Here, the base model denotes the stylization model equipped with \moduleD{} but without SAP or CR. 

As shown in Fig.~\ref{fig:abl}, the base model already preserves object structure reasonably well, but its stylization remains less faithful to the target style. 
When only CR is applied (\textit{w/o SAP}), the object semantics are better preserved, as indicated by the higher CLIP score compared with \textit{w/o CR}. 
This suggests that CR helps the model focus more on the object geometry by reducing the interference of color cues in the content image.
In contrast, when only SAP is applied (\textit{w/o CR}), the generated results exhibit better style alignment, especially in color appearance, which is consistent with the improved AdaIN metric. 
As shown in Fig.~\ref{fig:abl}, the base model and \textit{w/o SAP} remain less faithful to the target style, while \textit{w/o CR} and the full model produce more appropriate style colors. 

Finally, combining both SAP and CR leads to the best overall performance, showing that the two strategies play complementary roles in object preservation and style extraction.

\begin{figure}[t]
    \centering
    \includegraphics[width=1\linewidth]{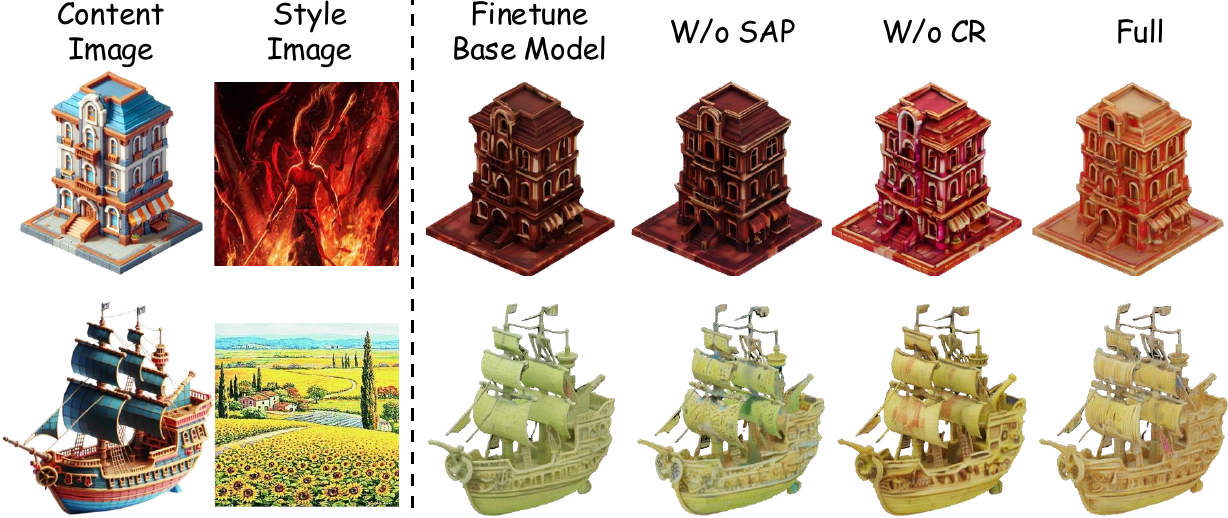}
    \caption{Ablation of Style Disentanglement Augmentation.}
    \label{fig:abl}
\end{figure}
\begin{table}[tp]
    \setlength\tabcolsep{6pt}
    \centering
    \small
    \caption{Ablation results of \moduleS{} on representative evaluation metrics.}
    \begin{tabular}{lcccc}
            \toprule
               & Finetune Base Model &  W/o SAP & W/o CR & Full \\ 
            \hline
            CLIP~$\uparrow$  & 79.488  &  79.869  &  79.822 & \textbf{80.096} \\
            AdaIN~$\downarrow$  & 91.948  & 91.879 &  91.748 & \textbf{91.682}\\
            \bottomrule
        \end{tabular}
    \label{tab:abl}
\end{table}

\section{Conclusions}
We propose a novel framework for generating stylized 3D assets.
By introducing \moduleD{} and \moduleS{}, we achieve explicit disentanglement and fusion of geometry and style features, enabling efficient style injection in the native 3D space.
We also develop an automated data generation pipeline and generate approximately 15K triplet samples, providing standardized supervision and benchmarks for model training and evaluation.
Experimental results demonstrate that our method excels in multiple 3D stylization benchmarks, qualitatively and quantitatively.

\begin{acks}
This research was supported by NSFC (NO. 62225604), Shenzhen Science and Technology Program (NO. JCYJ20240813114237048). This research was also supported by the Supercomputing Center of Nankai University (NKSC).
\end{acks}

\bibliographystyle{ACM-Reference-Format}
\balance
\bibliography{sample-base}

\clearpage

\appendix
\section*{Appendix}
This appendix is organized as follows:
\begin{enumerate}
    \item We provide additional analyses of \ourMthd{}, including ablations on DDCA and comparisons of different input-fusion strategies, as detailed in \secref{sec:analysis}.
    \item We describe the automated dataset curation pipeline, including the filtering of stylized 2D images and generated 3D assets, as detailed in \secref{sec:curation}.
    \item We present additional evaluations, including a comparison with external two-stage stylization pipelines and the user study, as detailed in \secref{sec:evaluation}.
    \item We provide additional qualitative results of stylized 3D assets across diverse objects and reference styles, as shown in \secref{sec:visualization}.
\end{enumerate}

\section{Additional Analysis of \ourMthd{}}\label{sec:analysis}

\subsection{Ablations on DDCA}

To further evaluate the design of the \moduleD{} (DDCA) module, we compare the full model with three variants. \textbf{MLP fusion} concatenates the DINO-based content features and CLIP-based style features and projects them into a unified conditioning representation using an MLP, which is then fed into a single cross-attention layer. \textbf{DINO fusion} retains the two-branch attention design but replaces the CLIP style features with DINO features, such that both branches are conditioned on DINO representations. \textbf{Style-only} removes the DINO-conditioned content-attention branch and retains only the CLIP-conditioned style-attention branch. All other components and experimental settings remain unchanged.

\begin{figure}[ht]
    \centering
    \includegraphics[width=\linewidth]{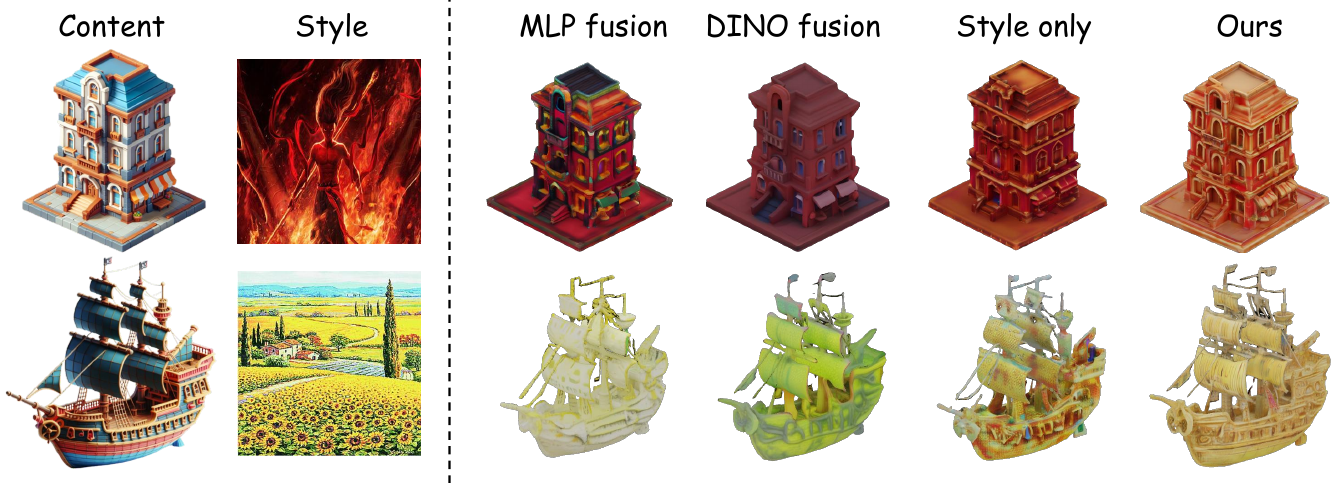}
    \caption{Qualitative ablation of DDCA variants. The full design achieves a better balance between structural preservation and style fidelity.}
    \label{fig:ddca_ablation}
\end{figure}

As shown in Fig.~\ref{fig:ddca_ablation}, the three variants reveal distinct failure modes. MLP fusion entangles geometry and style cues within a shared attention pathway, reducing control over their respective contributions. DINO fusion retains object-aware structural semantics but captures the reference style less effectively than CLIP-based conditioning. Style-only removes the content-conditioned attention branch, making latent refinement more susceptible to local structural deviations. In contrast, the full DDCA design uses separate DINO- and CLIP-conditioned attention branches and achieves a better balance between geometry preservation and style fidelity.

\subsection{Input Fusion Strategies}

\begin{figure*}
  \centering
   \includegraphics[width=0.95\linewidth]{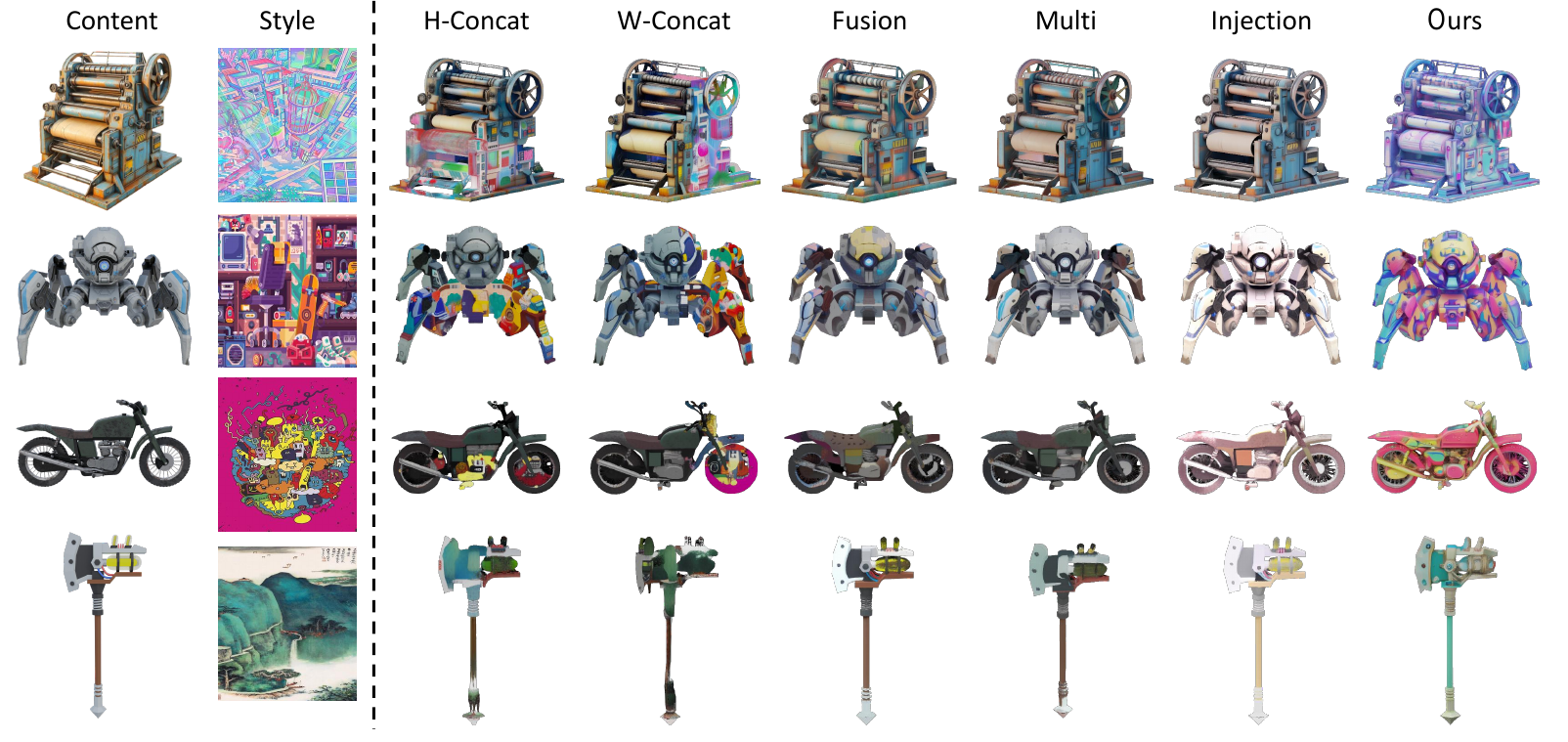}
   \caption{Comparison of different input fusion strategies under the TRELLIS pipeline.}
    \label{fig:concat}
\end{figure*}

\figref{fig:concat} provides additional analysis of the experimental setting in the main paper, where the content image and the style image are concatenated before being fed into TRELLIS.
Beyond simple spatial concatenation, we also explore multiple input fusion strategies, including concatenation along different axes, feature-level fusion, the official multi-image input setting, and statistic-based style injection.
These experiments are intended to verify whether controllable 3D stylization can be achieved by directly modifying the input format.
However, despite these extensive attempts, the resulting 3D assets remain unsatisfactory in terms of style controllability and visual consistency.
This suggests that input-level fusion alone is insufficient for controllable 3D stylization.

\section{Automated Dataset Curation Pipeline}\label{sec:curation}

To build a high-quality dataset consisting of stylized 3D assets and support the training and evaluation of 3D stylization models, we design an automated data curation pipeline consisting of two sequential stages. 
Specifically, the first stage filters stylized 2D images to ensure the reliability of both the style signals and the underlying content structure.
Then, the second stage performs filtering on the generated stylized 3D assets to ensure that each final 3D asset achieves sufficient visual quality and structural stability.

\myPara{Filtering of Stylized 2D Images.}
The first stage aims to select 2D stylized images with the most complete content structure and the highest style consistency from the candidate results generated by the OmniStyle model~\cite{wang2025omnistyle}. OmniStyle is pre-trained on a large-scale dataset of high-quality stylization triplets, enabling it to preserve the layout and semantic shape of the input content image while reliably injecting the target style during the stylization process. As a result, its generated candidates provide a reliable lower bound on content fidelity, forming a solid foundation for the subsequent filtering process.

For each content–style pair $I_c$ and $I_s$, we generate $K = 10$ stylized candidate images using different random seeds, denoted as $\{\hat{I}^k\}_{k=1}^K$. We then evaluate these candidates from two complementary perspectives: content preservation, measured by the Content Alignment Score (CAS)~\cite{xing2024csgo}, and style consistency, measured by Gram Matrix Similarity.

CAS is computed by performing an AdaIN-based de-stylization operation in the DINOv2 feature space~\cite{oquab2023dinov2}, which suppresses style attributes and emphasizes content-related structural information. It is defined as follows:
\begin{equation}
CAS(\hat{I}, I_c)
= \big\| \operatorname{Ada}(\phi(\hat{I})) - \operatorname{Ada}(\phi(I_c)) \big\|_2,
\end{equation}
where $\phi(\cdot)$ denotes the feature encoder, and the AdaIN de-stylization operator is defined as:
\begin{equation}
\operatorname{Ada}(F) = \frac{F - \mu(F)}{\sigma(F)},
\end{equation}
A lower CAS indicates a higher degree of geometric consistency between the candidate image and the content image.

Style consistency is assessed using Gram Matrix Similarity, which is based on the second-order statistics across feature channels extracted by VGG-19~\cite{simonyan2014very}. For a feature map $F \in \mathbb{R}^{C \times H \times W}$, the Gram matrix is defined as:
\begin{equation}
G = \frac{1}{C \cdot H \cdot W} FF^\top,
\end{equation}
The style difference between the candidate image and the style image is measured by the Frobenius norm, defined as:
\begin{equation}
L_{\text{Gram}} = \|G_{\text{ref}} - G_{\text{gen}}\|_{F},
\end{equation}
A smaller style difference indicates that the image’s style is closer to the target style.

To combine the two metrics on a unified scale, we first apply min–max normalization to each of them and then compute a unified aggregated score as follows:
\begin{equation}
\begin{gathered}
\widehat{CAS}^k = \frac{CAS^k - CAS_{\min}}{CAS_{\max} - CAS_{\min}}, \\[3pt]
\widehat{L}_{\text{Gram}}^k = \frac{L_{\text{Gram}}^k - L_{\min}}{L_{\max} - L_{\min}}, \\[3pt]
Q^k = \alpha\,\widehat{CAS}^k + (1-\alpha)\,\widehat{L}_{\text{Gram}}^k ,
\end{gathered}
\end{equation}
where $\alpha = 0.3$. Since both normalized metrics are defined such that smaller deviations indicate better performance, we select the candidate image with the lowest aggregated score $Q^k$ among the ten candidates as the final stylized 2D result.

This stage of filtering ensures that the inputs to the subsequent 3D stylization process meet reliable standards in both style consistency and content fidelity.

\myPara{Filtering of Stylized 3D Assets.}
After obtaining the filtered 2D style images, we generate their corresponding 3D stylized assets using the Hunyuan3D-2.1~\cite{hunyuan3d2025hunyuan3d} model, in which the content image is fed into the geometry generation stage and the filtered stylized image is used in the texture generation stage. 
We then further perform filtering on these generated 3D stylized assets to ensure that the final dataset contains 3D models with stable visual performance under rendering.
Following the multi-view quality assessment strategy used in TRELLIS~\cite{xiang2025structured}, we render each 3D asset from four fixed views, resulting in a set of images $\{R_i\}_{i=1}^{4}$. 
Next, we evaluate the rendering quality from two complementary perspectives: the aesthetic score of the rendered images and the CLIP-based~\cite{radford2021learning} image similarity between the renderings and the style reference.

The rendering aesthetic score is computed using the image-level aesthetic predictor adopted in TRELLIS~\cite{xiang2025structured}, which provides a general measure of the visual plausibility of rendered images. However, because this predictor is trained on large-scale natural image preference datasets rather than stylized visual domains, stylized renderings may occasionally receive lower aesthetic scores even when their 3D geometry and style transfer are correct. Consequently, aesthetic score alone is insufficient for filtering stylized 3D assets. To avoid incorrectly discarding valid stylized results, we complement it with the CLIP-based similarity between the rendered images and the style reference. A higher CLIP similarity indicates stronger preservation of the intended stylistic semantics, helping distinguish true rendering failures from style-driven aesthetic deviations.

\begin{figure*}[ht]
  \centering
  \includegraphics[width=0.9\linewidth]{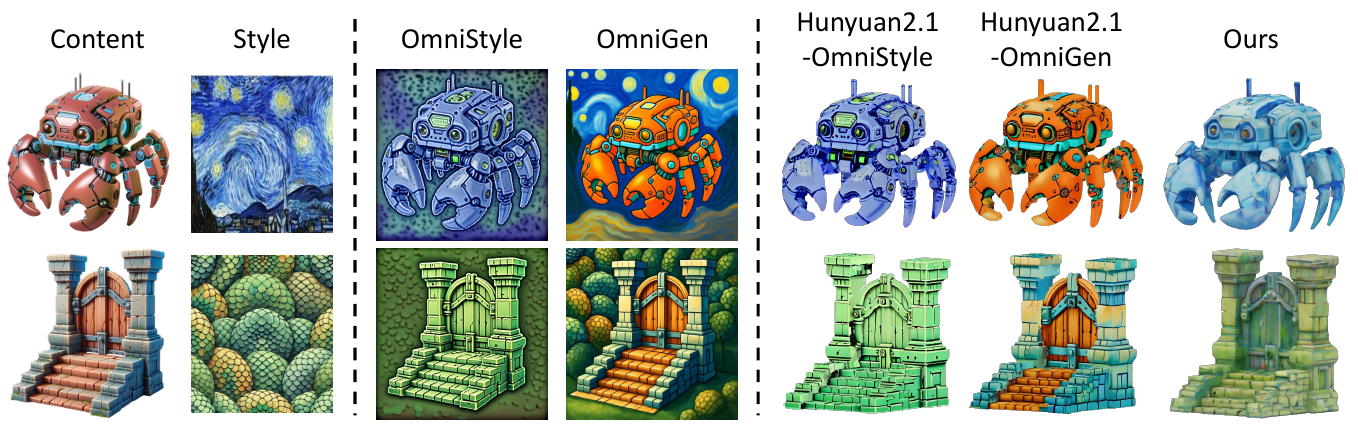}
  \caption{Discussion of our method and a two-stage stylization pipeline.}
  \label{fig:twostage}
\end{figure*}

We compute the average aesthetic score and the average CLIP similarity across the four rendered views, denoted as $S_{\text{aesthetic}}$ and $S_{\text{CLIP}}$, respectively. We then filter out low-quality assets using the following dual-criterion condition:
\begin{equation}
S_{\text{aesthetic}} < 4.5 \quad \text{and} \quad S_{\text{CLIP}} < 0.6.
\end{equation}
An asset is discarded only when it performs poorly in both dimensions—rendering quality and preservation of stylistic semantics—simultaneously. In practice, approximately $ 10\% $ of the initially generated 3D assets are discarded during this stage, resulting in a final curated dataset of about 15K stylized 3D triplets.

Through this complementary two-stage filtration, we construct a high-quality dataset of stylized 3D assets with strong style consistency, high geometric fidelity, and stable rendering quality. This dataset provides a robust and reliable foundation for the training and evaluation of our \ourMthd{} and future 3D stylization methods.

\section{Additional Evaluation}\label{sec:evaluation}

\begin{figure}[ht]
    \centering
    \includegraphics[width=1\linewidth]{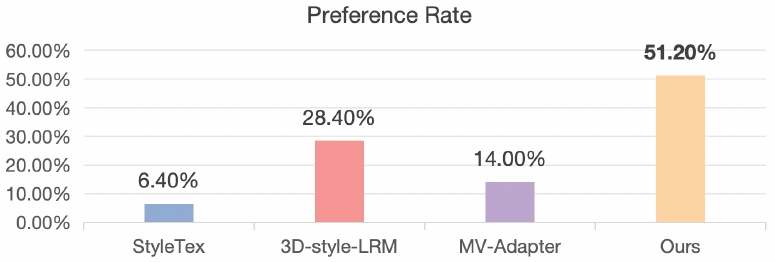}
    \caption{User study results in terms of preference rate.}
    \label{fig:user}
\end{figure}

\subsection{Comparison with Two-Stage Pipelines}

We further discuss the difference between our native 3D end-to-end method and a two-stage pipeline that first performs 2D style transfer and then generates a 3D asset.
Specifically, given a content image and a style image, we first feed them into a 2D stylization model to obtain a stylized image. In our experiments, we use OmniStyle~\cite{wang2025omnistyle} and OmniGen2~\cite{wu2025omnigen2} as representative examples.
We then use Hunyuan3D-2.1~\cite{hunyuan3d2025hunyuan3d} as the downstream 3D generator: the original content image is used as the input to its first-stage geometry generation, while the stylized image is used as the input to its second-stage texture generation, finally producing a stylized 3D asset.
In this way, the 3D stylization task is decomposed into 2D stylization followed by 3D generation.
In principle, if the intermediate stylized image is highly accurate, the subsequent 3D generation problem becomes considerably easier, which may lead to strong final results.
This is also why we adopt a similar two-stage pipeline in our dataset construction.

However, it is important to note that our dataset pipeline is an offline curation process, where we carefully filter the outputs of both the 2D stylization stage and the 3D generation stage to ensure data quality.
This does not contradict our discussion here, since the two-stage pipeline is used there as an offline data production process with strict filtering, rather than as a direct test-time generation solution.
In contrast, when used as a direct generation pipeline, the two-stage formulation is more prone to error accumulation.
As shown in \figref{fig:twostage}, existing 2D stylization methods may produce biased intermediate results, and in some cases mainly stylize the background instead of the target object.
Such deviations are then passed to the subsequent 3D generation stage, causing the final 3D asset to differ substantially from the desired style reference.
Moreover, the two-stage pipeline requires an additional stylization model and an extra inference stage, making it slower and more computationally demanding than our native 3D end-to-end method.

\subsection{User Study}

We conducted a user study with 50 participants, all of whom were graduate students or researchers in computer vision or graphics and were familiar with 3D asset evaluation. Each participant evaluated five representative cases covering diverse object categories and style types.

For each case, participants were shown the content image, the style image, and rendered results from StyleTex, 3D-style-LRM, MV-Adapter, and \ourMthd{}. They were asked to select the result that best balanced style alignment, structural preservation, and overall visual quality. We aggregated all selections to compute the preference rate of each method.

As shown in Fig.~\ref{fig:user}, \ourMthd{} achieves the highest preference rate of 51.2\%. This ranking is consistent with the quantitative and qualitative evaluations in the main paper.

\section{More Visualization Results}\label{sec:visualization}
\begin{figure*}
    \centering
    \includegraphics[width=0.95\linewidth]{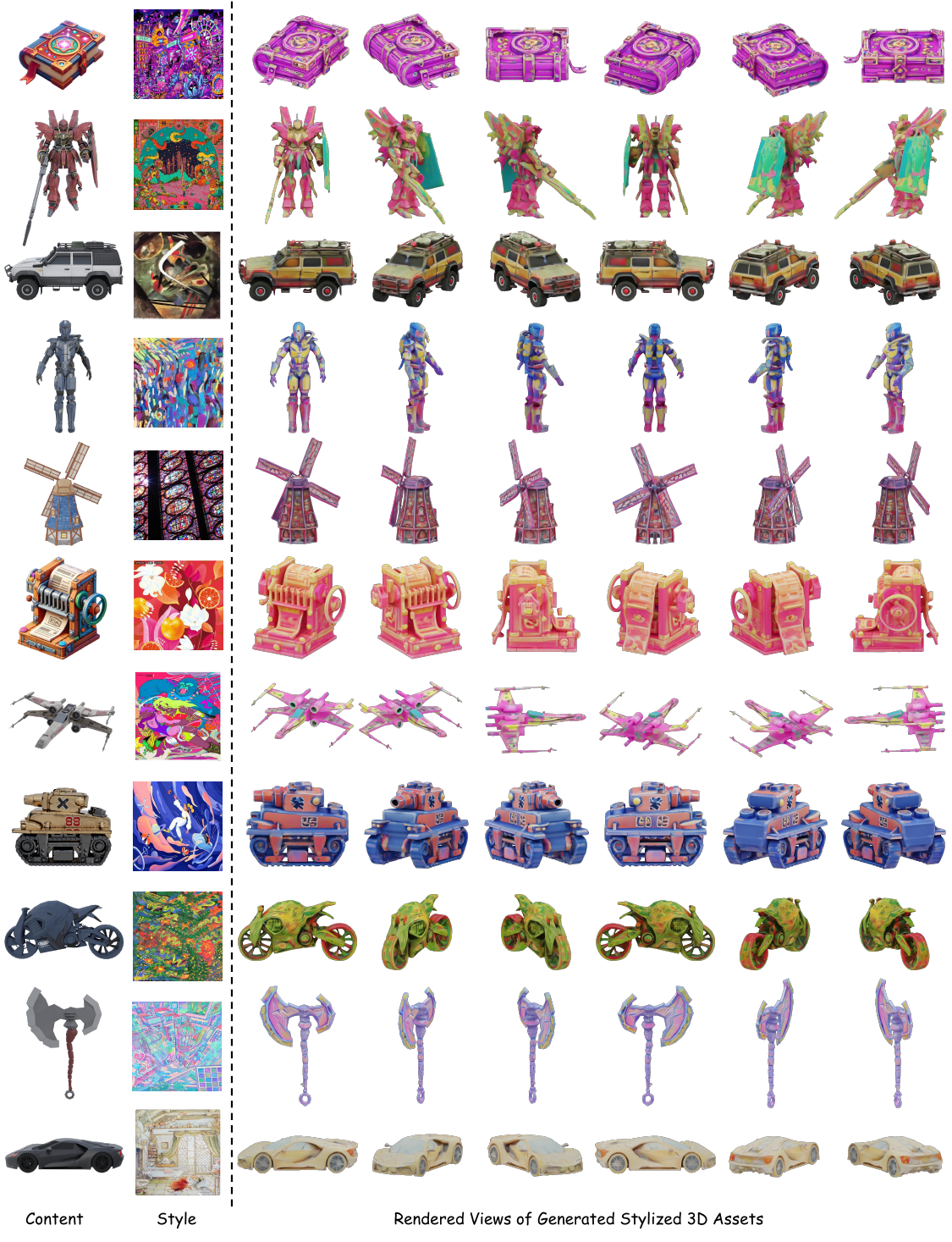}
    \caption{
    More visualization results of the 3D stylized assets from our \ourMthd{}, featuring multi-view renderings of various object types and style references.
    It can be observed that our method achieves high visual quality and texture integrity across different views.
    }
    \label{fig:supp}
\end{figure*}

\figref{fig:supp} presents additional 3D stylized assets generated by our method, further validating its adaptability and effectiveness in diverse scenes. Through multi-view renderings, the figure showcases various object types fused with different style references. Each set of images demonstrates the combination of the content and style images, illustrating that our method accurately injects style into the generated 3D assets while maintaining geometric consistency, resulting in strong style consistency and high visual quality. Additionally, the rendering views in the figure cover a 360-degree rotation of the objects, clearly showing the geometric details and style features from all angles after style injection.

\end{document}